\documentclass[letterpaper, 10 pt, conference]{ieeeconf}  % Comment this line out if you need a4paper

\IEEEoverridecommandlockouts                         
\usepackage{amsmath,graphicx,booktabs}
\usepackage{amssymb}
\usepackage[pagebackref=true,breaklinks=true,letterpaper=true,colorlinks,bookmarks=false]{hyperref}
\usepackage{adjustbox}
\usepackage{xcolor}
\usepackage{float}
\usepackage{fancyhdr}
\usepackage{mathptmx}

\usepackage[linesnumbered,ruled,boxed,norelsize,vlined,commentsnumbered]{algorithm2e}

\usepackage{array} 
\begin{document}

\title{SQuaD-SQL: Efficient Text-to-SQL with Small Language Models via LLM-Guided Knowledge Distillation}

\author{Wangyu Wu$^{1,5}$, Xiaojian Lin$^{4}$, Rong Fu$^{2}$, Zaiyang Yu$^{3}$, Xuhang Chen$^{7}$, Wenjun Yu$^{6}$, Zhenhong Chen$^{5^{*}}$ \\
\\
$^{1}$The University of Liverpool \quad $^{2}$University of Macau \\ 
\quad $^{3}$University of Chinese Academy of Sciences $^{4}$Tsinghua University \quad $^{5}$ Microsoft \quad 
\\ $^{6}$ Shanghai University of International Business and Economics \quad $^{7}$ Huizhou University \quad 
\thanks{$^{*}$Corresponding authors: daisychen37@foxmail.com}% <-this % stops a space
\thanks{}}

\maketitle

\begin{abstract}
Text-to-SQL is a fundamental task in natural language processing that enables users to interact with structured databases using natural language. While large language models (LLMs) have demonstrated remarkable performance on this task, their substantial computational requirements hinder deployment in resource-constrained settings. In this paper, we introduce \textbf{SQuaD-SQL} (\textit{Small-Qualified and Distilled for SQL}), a novel approach that empowers small language models (SLMs) to approach the performance of LLMs on the Text-to-SQL task while significantly improving efficiency through knowledge distillation and synthetic data generation. Our method comprises three key components: (1) \textit{LLM-based synthetic data generation}, where structured knowledge is extracted from LLMs via carefully designed prompting strategies; (2) \textit{parameter-efficient fine-tuning}, enabling full model training on a single consumer-grade GPU; and (3) \textit{domain-adaptive fine-tuning}, where domain-specific synthetic data further enhances performance in targeted domains. Experiments on the WikiSQL dataset demonstrate that SQuaD-SQL achieve execution accuracies of 86.9\% in Test data, respectively, approaching the performance of LLMs, while offering faster inference and lower memory usage. These results suggest that, with proper training strategies, SLMs can serve as practical and efficient alternatives for Text-to-SQL applications in resource-limited environments.

\end{abstract}

\section{Introduction}

Text-to-SQL enables users to query structured databases using natural language, significantly lowering the barrier for database access. Early rule- or template-based systems lacked scalability, while neural models such as Seq2SQL~\cite{zhong2017seq2sql} and SQLNet~\cite{xu2017sqlnet} improved performance via sketch-based decoding and reinforcement learning. Recent advances in natural language interfaces to structured data, particularly Text-to-SQL, have largely followed a scale-driven paradigm. Text-to-SQL enables users to query structured databases using natural language, significantly lowering the barrier for database access. Early rule- or template-based systems lacked scalability, while neural models such as Seq2SQL~\cite{zhong2017seq2sql} and SQLNet~\cite{xu2017sqlnet} improved performance via sketch-based decoding and reinforcement learning. Subsequent work introduced more complex datasets such as Spider~\cite{yu2018spider} and structure-aware architectures like RAT-SQL~\cite{wang2020rat}. Pretrained tabular models, including TAPAS~\cite{herzig2020tapas} and TaBERT~\cite{yin2020tabert}, further advanced performance by incorporating schema and table representations.

\begin{figure}[t]
\centering
\includegraphics[width=0.9\columnwidth]{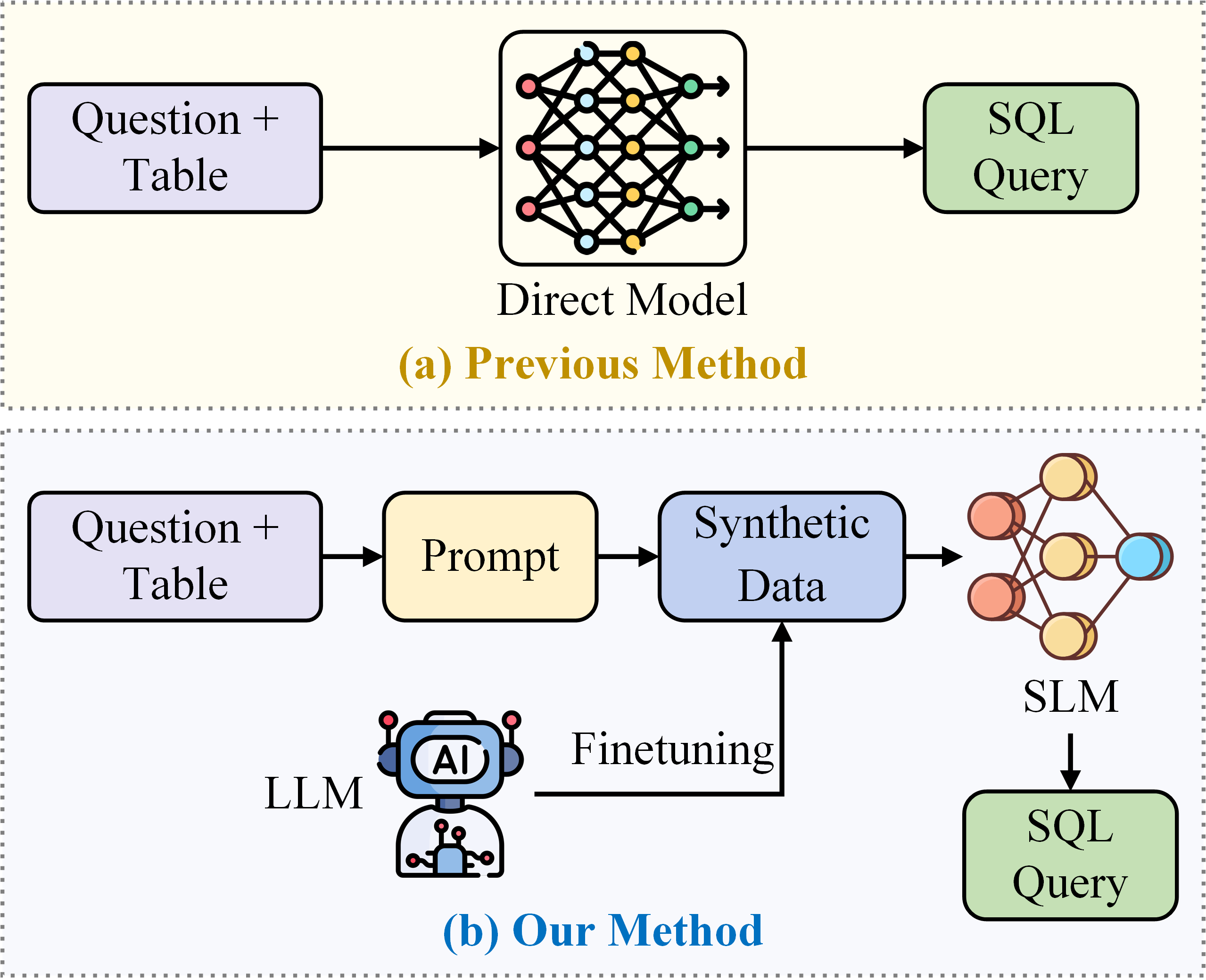}
\caption{Overview of the proposed teacher--student learning framework. A large language model provides structured instructional signals that guide a compact student model to internalize Text-to-SQL reasoning patterns without relying on manual annotation.}
\label{fig:main}
\vspace{-2em}
\end{figure}

More recently, large language models (LLMs)~\cite{brown2020language, scholak2021picard} have demonstrated strong reasoning capabilities on Text-to-SQL tasks, often achieving state-of-the-art performance. However, these gains come at the cost of substantial computational resources. LLM-based systems~\cite{wu2026llm,wu2025image,wu2025contrastive,cai2024relation} typically suffer from high inference latency, expensive deployment requirements involving multiple high-end GPUs, and resource-intensive training and fine-tuning procedures. Such demands diverge sharply from the efficiency observed in human learning and significantly limit the practical deployment of Text-to-SQL systems in real-world, resource-constrained settings.

This contrast raises a fundamental question from a cognitive perspective: \emph{can compact models acquire structured reasoning abilities comparable to large models when guided by appropriate instructional signals, rather than raw scale alone?} Small language models (SLMs) offer clear advantages in efficiency and accessibility, yet training them from scratch often leads to severe performance degradation on complex reasoning tasks such as Text-to-SQL. Bridging the gap between efficiency and reasoning competence remains a key challenge. In this work, we propose \textbf{SQuaD-SQL}, a teacher--student learning framework that enables small language models to internalize structured SQL reasoning behaviors under resource constraints. Drawing inspiration from instructional learning in humans, we treat large language models as teachers that generate structured supervision signals, while compact student models learn to abstract and internalize these reasoning patterns without requiring the teacher at inference time.

Specifically, we design customized prompt templates tailored to the Text-to-SQL task to elicit high-quality, structured instructional signals from the teacher model. These signals serve as synthetic supervision that guides the student model toward learning compositional SQL reasoning. To ensure efficiency, we adopt parameter-efficient fine-tuning techniques such as LoRA, allowing the entire training process to be conducted on a single consumer-grade GPU (e.g., RTX 4090). Furthermore, we enhance domain adaptability by generating targeted synthetic SQL examples for specific application domains, enabling the student model to generalize more effectively across diverse settings.

We evaluate our approach on the WikiSQL benchmark. Experimental results show that a 1.5B-parameter student model (Qwen-1.5B) achieves execution accuracy comparable to substantially larger models while offering significantly improved inference efficiency. These findings demonstrate that structured reasoning can be effectively learned under resource constraints, providing a computational account of how complex symbolic behaviors may emerge through guided instruction rather than scale alone.

\textbf{As illustrated in Figure~\ref{fig:main}, our main contributions are summarized as follows:}
\begin{itemize}
    \item We introduce a teacher--student learning framework that models structured reasoning acquisition under resource constraints by transferring instructional signals from large language models to compact students.
    \item We propose SQuaD-SQL, a lightweight and efficient training approach for Text-to-SQL based on parameter-efficient fine-tuning, enabling practical deployment using limited computational resources.
    \item We conduct comprehensive experiments on WikiSQL, demonstrating that our approach achieves strong reasoning performance while maintaining substantial efficiency advantages over LLM-based baselines.
\end{itemize}

\section{Related Work}

\subsection{Text-to-SQL Task}

Text-to-SQL is a fundamental NLP task that translates natural language questions into executable SQL queries, enabling non-experts to access databases. Early rule-based methods~\cite{li2014constructing, iyer2017learning} suffered from limited generalization. With deep learning, neural models like Seq2SQL~\cite{zhong2017seq2sql} and SQLNet~\cite{xu2017sqlnet} improved accuracy on benchmarks like WikiSQL~\cite{zhong2017seq2sql} using attention and sketch-based decoding.

More complex datasets like Spider~\cite{yu2018spider} required models that generalize to unseen schemas. Structure-aware models like RAT-SQL~\cite{wang2020rat} used relation-aware attention, while IRNet and BRIDGE enhanced schema linking. Pretrained language models (PLMs) further advanced performance: TAPAS~\cite{herzig2020tapas} and TaBERT~\cite{yin2020tabert} learned table-text representations, and PICARD~\cite{scholak2021picard} enforced syntactic validity via constrained decoding.

However, state-of-the-art methods rely on large models (e.g., T5 or Codex), which are computationally expensive. Our work addresses this by exploring knowledge distillation and efficient finetuning to enable small language models (SLMs) for Text-to-SQL.

\begin{figure*}[t]
\centering
\includegraphics[width=1\linewidth]{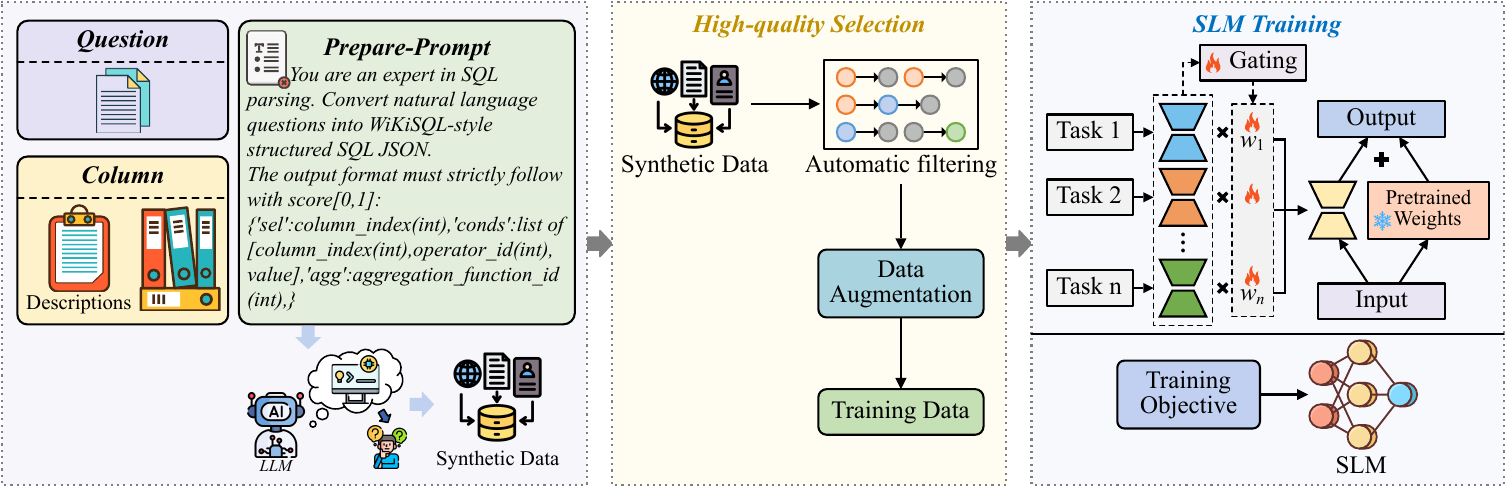}
\caption{Method overview: Using LLMs to generate synthetic data, training small language models, and performing domain-adaptive fine-tuning.}
\label{fig:method_overview}
\end{figure*}
\subsection{Language Models}

With the emergence of large language models such as GPT-4o~\cite{brown2020language}, PaLM~\cite{chowdhery2022palm}, and LLaMA~\cite{touvron2023llama}, the performance of the Text-to-SQL task has been further improved. These models acquire strong language understanding and generation capabilities through large-scale pretraining, enabling them to directly comprehend user intent from natural language instructions and generate corresponding SQL queries.

Large models specifically optimized for code generation tasks, such as SQLCoder~\cite{sqlcoder2023} and CodeLlama~\cite{roziere2023code}, perform particularly well on Text-to-SQL tasks. By pretraining on code corpora, they develop a deep understanding of SQL syntax and database operations. However, such large models typically require billions or even hundreds of billions of parameters, resulting in high inference latency and deployment costs, making them difficult to use in resource-constrained environments.

To address the resource demands of large models, researchers have started to explore the potential of small language models. Models like the Qwen series~\cite{qwen2023}, Phi~\cite{gunasekar2023textbooks}, and TinyLlama~\cite{tinyllama2023} adopt carefully designed pretraining strategies to maintain competitive performance with significantly fewer parameters.

Knowledge distillation, first proposed by Hinton et al.~\cite{hinton2015distilling}, is an effective method for improving the performance of small models. In NLP, works like DistilBERT~\cite{sanh2019distilbert} and TinyBERT~\cite{jiao2020tinybert} successfully compressed the BERT model while preserving most of its performance. Recently, researchers have applied distillation techniques to large language models. Works such as Alpaca~\cite{alpaca2023} and Vicuna~\cite{vicuna2023} have enhanced instruction-following capabilities in small models by distilling knowledge from large models.

\subsection{Synthetic Data Generation}

Synthetic data generation is an effective solution to the problem of data scarcity. In the Text-to-SQL domain, early works such as~\cite{yu2018typesql, guo2019towards} used templates and rules to generate synthetic queries for data augmentation. With the development of large language models, researchers have begun exploring the use of these models to generate high-quality synthetic data. Works such as Self-Instruct~\cite{wang2022self} and WizardLM~\cite{xu2023wizardlm} show that large models can generate diverse instruction data via self-guided prompting, which can then be used to train smaller models.

In Text-to-SQL tasks, methods like SQLGen~\cite{sqlgen2023} and DB-GPT~\cite{dbgpt2023} explore generating SQL-natural language pairs using large models to enhance generalization. However, these efforts primarily focus on fine-tuning large models, with limited exploration on how to effectively use synthetic data to train resource-efficient small models.

Our work is related to the above but focuses specifically on leveraging knowledge distillation and synthetic data generation to enable small language models (such as Qwen-1.5B) to achieve near-large-model performance on Text-to-SQL tasks, while maintaining significant efficiency advantages. Our method not only emphasizes model performance but also highlights training and inference efficiency, enabling the entire process to be completed on a single consumer-grade GPU.

\section{Method}

This section details our proposed method, which combines knowledge distillation and synthetic data generation to enable small language models to achieve near-large-model performance on the Text-to-SQL task while maintaining significant efficiency advantages. 

\subsection{Overview}

The goal of this work is to enable SLMs to perform competitively on the Text-to-SQL task while maintaining efficiency suitable for deployment in resource-constrained environments. To achieve this, we propose a framework that leverages large language models (LLMs) for synthetic data generation, applies rigorous data quality control to ensure high-quality training samples, and employs lightweight training strategies such as LoRA to efficiently fine-tune SLMs. Figure~\ref{fig:method_overview} provides an overview of our proposed method, which consists of three main stages: LLM-based data generation, high-quality data selection, and LoRA-based SLM training.

\subsection{LLM-Based Data Generation.}
To reduce the training cost of Text-to-SQL models while preserving performance, we leverage large language models (LLMs), such as GPT-4o, as data generators to produce high-quality supervision in WikiSQL format for small language models (SLMs). Rather than relying on LLMs for inference, we use them to synthesize paired natural language questions and structured SQL labels.

We extract schema information from the WikiSQL dataset, including column names and optionally sampled rows, and use this information to construct instruction-style prompts. These prompts guide the LLM to generate structured SQL outputs in a consistent and syntactically correct format. The generated data covers a wide range of question complexities and logical conditions by varying schema contexts and query types.

The resulting synthetic pairs provide a diverse and robust training signal for downstream fine-tuning. We formalize the data generation process using a teacher-student formulation:
\begin{equation}
(x_i, y_i) \leftarrow f_{\text{LLM}}(\text{prompt}_i),
\end{equation}
where $x_i$ is the generated natural language question and $y_i$ is the corresponding SQL query, generated by the teacher model $f_{\text{LLM}}$ from input prompt $\text{prompt}_i$.

\subsection{High-Quality Selection}

Given the synthetic SQL labels generated by large language models (LLMs) for the WikiSQL dataset, it is critical to ensure that only high-quality examples are retained for training small language models (SLMs). Since LLMs may occasionally generate incorrect, incomplete, or overly complex SQL statements, we design a multi-stage filtering pipeline that aims to select pseudo-labels with high precision.

The first step in our filtering pipeline is to verify the syntactic validity of the SQL statement. For each generated pair $(x_i, y_i)$, where $x_i$ is a natural language query and $y_i$ is the corresponding SQL, we apply a SQL parser to ensure that $y_i$ conforms to the SQL grammar rules. Examples that fail to parse are immediately discarded.

Next, we introduce an LLM-based self-evaluation scoring mechanism. For each SQL query $y_i$, we prompt the same LLM that generated the query to assess the quality of the SQL based on its relevance to the input $x_i$, correctness with respect to the schema, and completeness of the query. The LLM returns a confidence score $s_i \in [0,1]$ for each example. We define a threshold $\lambda$ and retain only examples where $s_i > \lambda$.

In addition to syntactic and semantic scoring, we conduct rule-based execution validation to verify logical soundness. Each SQL query $y_i$ is executed on the corresponding table in the WikiSQL database to check whether the query produces valid and non-empty results. This step helps eliminate logically invalid SQL that may pass syntax checks but are semantically meaningless or overly generic.

After applying the above filtering mechanisms, we collect the resulting subset $\mathcal{D}_{\text{filtered}} = \{(x_i, y_i)\}_{i=1}^{N'}$, which we treat as high-confidence pseudo-labeled data. These examples are then used to fine-tune SLMs, enabling them to inherit the structured reasoning capabilities of LLMs through indirect supervision.

Formally, let $f_{\text{LLM}}$ be the generation model and $f_{\text{score}}$ be the evaluation model (which may be the same as $f_{\text{LLM}}$). The filtering process is summarized by the following formulation:

\begin{align}
\mathcal{D}_{\text{filtered}} = \Big\{(x_i, y_i) \ \Big| \ 
& y_i = f_{\text{LLM}}(x_i), \nonumber \\
& \text{is\_valid}(y_i) = 1, \nonumber \\
& s_i = f_{\text{score}}(x_i, y_i), \nonumber \\
& s_i > \lambda \Big\}
\end{align}

To better illustrate the filtering pipeline, we present the procedure in pseudocode:

\begin{algorithm}
\caption{High-Quality Selection of \\LLM-Generated Data}
\KwIn{Natural language queries $\{x_1, \dots, x_N\}$, schema $\mathcal{S}$, LLM $f_{\text{LLM}}$, scorer $f_{\text{score}}$, threshold $\lambda$}
\KwOut{Filtered dataset $\mathcal{D}_{\text{filtered}}$}
$\mathcal{D}_{\text{filtered}} \leftarrow \emptyset$ \\
\ForEach{$x_i \in \{x_1, \dots, x_N\}$}{
    $y_i \leftarrow f_{\text{LLM}}(x_i, \mathcal{S})$ \\
    \If{\texttt{not} \textsc{IsValidSQL}($y_i$)}{
        \texttt{continue}
    }
    $s_i \leftarrow f_{\text{score}}(x_i, y_i, \mathcal{S})$ \\
    \If{$s_i \le \lambda$}{
        \texttt{continue}
    }
    \If{\texttt{not} \textsc{ExecValid}($y_i$)}{
        \texttt{continue}
    }
    $\mathcal{D}_{\text{filtered}} \leftarrow \mathcal{D}_{\text{filtered}} \cup \{(x_i, y_i)\}$
}
\Return $\mathcal{D}_{\text{filtered}}$
\end{algorithm}

This pipeline ensures that the pseudo-labeled training data used for SLM supervision exhibits both structural validity and semantic alignment with the natural language input. The LLM-based scoring step provides a fine-grained self-assessment of label quality, while the rule-based checks serve as complementary safeguards. Empirically, we find that applying this three-stage filter significantly improves the downstream training stability and final execution accuracy of SLMs on the Text-to-SQL task.

\subsection{Training SLM with Synthetic Data and LoRA Adaptation}

To leverage the high-quality synthetic data generated and filtered in previous steps, we fine-tune a SLM using the WikiSQL task as the target benchmark. We employ Qwen-1.5B as the base model and integrate Low-Rank Adaptation (LoRA) to enable efficient fine-tuning. This section outlines the training architecture, LoRA injection strategy, and the loss functions used in optimization.

\paragraph{Model Architecture.} 
Qwen-0.5B is a decoder-only transformer model with standard causal language modeling architecture. Each transformer block applies multi-head self-attention and feedforward layers. The attention mechanism computes attention scores using the standard key-query-value (KQV) formulation:
\begin{equation}
\text{Attention}(Q, K, V) = \text{softmax} \left( \frac{QK^T}{\sqrt{d_k}} \right) V,
\end{equation}
where $Q = XW^Q$, $K = XW^K$, $V = XW^V$ are the query, key, and value projections of the input sequence $X$, and $d_k$ is the dimensionality of the key vectors.
\begin{figure}[t]
\centering
\includegraphics[width=0.75\columnwidth]{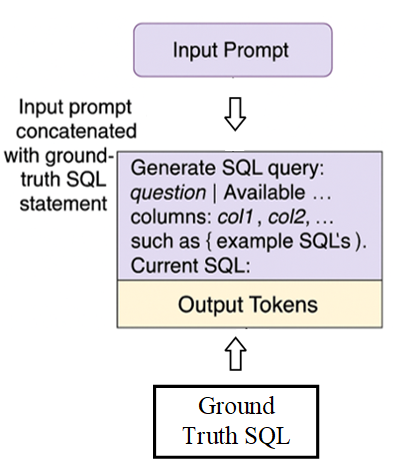}
\caption{Masked causal language modeling loss applied to SQL generation. The input prompt is concatenated with ground-truth SQL, and loss is computed only over the SQL tokens..}
\label{fig:loss}
\end{figure}

\paragraph{LoRA Integration.} 
To reduce the number of trainable parameters while maintaining performance, we adopt LoRA~\cite{hu2021lora}. LoRA introduces trainable low-rank decomposition matrices $(A, B)$ into the attention projections, such that:
\begin{equation}
W^Q_{\text{LoRA}} = W^Q + \alpha BA, \quad \text{where } A \in \mathbb{R}^{r \times d}, B \in \mathbb{R}^{d \times r},
\end{equation}
and $\alpha$ is a scaling factor, typically set to $\frac{1}{r}$. We apply LoRA to the attention layers' query and value projections ($q\_proj$, $v\_proj$), enabling efficient adaptation using a small number of parameters.
\paragraph{Training Setup.}
We fine-tune the model using high-quality synthetic SQL generation data formatted as:
\begin{quote}
\texttt{Generate SQL query: \textit{question} | Available columns: \textit{col1, col2, ...} such as \{\textit{example SQLs}\}. Current SQL:}
\end{quote}
As shown in Figure~\ref{fig:loss}, the input prompt is concatenated with the ground-truth SQL statement to form a single input sequence. During training, we adopt a causal language modeling loss, but compute the loss only over the SQL generation portion by masking the prompt tokens. This is implemented via label masking using \texttt{-100} in the label tensor for the prompt segment:
\begin{equation}
\mathcal{L}_{\text{SQL}} = - \sum_{t \in \text{target}} \log P(y_t \mid y_{<t}, x),
\end{equation}
where $x$ denotes the prompt tokens, and $y_t$ is the $t$-th token in the SQL output. This masked causal objective enables the model to focus on learning the SQL generation behavior without being penalized for the prompt portion.

\paragraph{Overall Objective.}
The final loss function during fine-tuning aggregates the masked causal language modeling loss across all training samples:
\begin{equation}
\mathcal{L}_{\text{total}} = \frac{1}{N} \sum_{i=1}^{N} \mathcal{L}_{\text{SQL}}^{(i)},
\end{equation}
where $N$ is the number of samples in a mini-batch. We use the AdamW optimizer with a linear learning rate scheduler and warm-up steps. This efficient fine-tuning strategy, combined with synthetic supervision, enables the SLM to effectively acquire SQL generation capabilities with limited resources.

\section{Experiment}

This section details our experimental settings, evaluation metrics, and results, validating the effectiveness and efficiency of our proposed method on the Text-to-SQL task.

\subsection{Experimental Setup}

\subsubsection{Dataset and Baselines}

We primarily conduct experiments on the WikiSQL dataset, a large-scale benchmark widely used in Text-to-SQL tasks. The dataset contains 80,654 natural language question-SQL query pairs, covering 24,241 unique tables. Following the standard split, the dataset is divided into 56,355 training samples, 8,421 validation samples, and 15,878 test samples. A key characteristic of WikiSQL is that each query is constructed based on a single table, with SQL operations including \texttt{SELECT}, \texttt{WHERE}, and \texttt{GROUP BY}. Adhering to the standard data partitioning, we use 56,355 samples for training, 8,421 for validation, and 15,878 for testing.

In comparative experiments, we evaluate our method against three categories of baselines. The first category consists of large language models (LLMs), such as GPT-4o. The second category includes smaller language models, specifically Qwen-0.5B and Qwen-1.5B directly queried on WikiSQL. The final category comprises traditional Text-to-SQL models, such as the slot-filling-based SQLNet and BRIDGE, which enhances semantic understanding by integrating pre-trained language models.

%========== LoRA  Rank  Line Plot ==========
\begin{figure*}[t]
\centering
\includegraphics[width=0.8\linewidth]{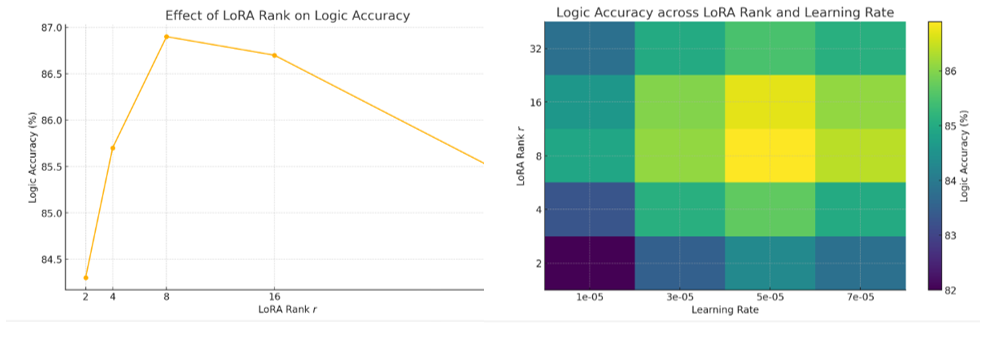} % 将文件保存到 figs/ 后引用
  \caption{Impact of LoRA rank $r$ on execution accuracy. Peak performance (86.9\%) is achieved at $r{=}8$, with diminishing returns beyond this rank. And Heatmap of execution accuracy across LoRA ranks ($r$) and learning rates.}
  \label{fig:lora_rank_line}
\end{figure*}

\subsection{Implementation Details}

All experiments are conducted on a single NVIDIA RTX 4090 GPU with 24GB of memory. Using GPT-4o, we synthesize approximately 50,000 high-quality natural language-SQL query pairs that adhere to the WikiSQL data structure format. The task involves converting natural language questions into corresponding SQL queries based on table schema information. Each sample in WikiSQL contains a table and a structured SQL query in the format \texttt{\{agg, sel, conds\}}.

We implement the training pipeline using HuggingFace Transformers. For parameter-efficient fine-tuning, we employ Low-Rank Adaptation (LoRA) with rank $ r = 8 $ and a scaling factor $ \alpha = 32 $ applied to the \texttt{q\_proj} and \texttt{v\_proj} modules. Mixed-precision FP16 training is used to accelerate training and reduce memory consumption. The model is trained for 10 epochs with a batch size of 4 and a learning rate of 5e-5. During training, the base model is frozen, and only the LoRA-injected layers are updated.

Each training sample's input consists of a natural language question and its corresponding table column information, supplemented with JSON-formatted SQL annotation examples. The target output is a standard SQL query string. During tokenization, we apply loss masking to the output portion to ensure the model's causal decoding capability. For evaluation, we report the following two metrics: (1) \textit{Execution Accuracy}: whether the predicted SQL produces the same execution result as the reference query; and (2) \textit{Logical Form Accuracy}: whether the predicted SQL query is logically equivalent to the reference answer.

\subsection{Performance Comparison}

Table~\ref{tab:main_results} reports the logical-form execution accuracy of our method and representative baselines on WikiSQL. We group the baselines into: (i) a strong large language model (GPT-4o), (ii) classic and recent Text-to-SQL or structured QA models, (iii) off-the-shelf small language models (SLMs) without our synthetic-data training, and (iv) our proposed SQuaD-SQL trained on GPT-4o-generated synthetic supervision. Among all models, GPT-4o achieves the best performance (92.2\% dev / 93.5\% test), demonstrating the advantage of large-scale pretraining and strong in-context reasoning. However, its inference cost and deployment requirements remain prohibitive for many resource-constrained scenarios.

Standard Text-to-SQL baselines achieve moderate results. Early neural parsers such as SQLNet reach 69.8\%/72.1\%, while reinforcement-learning-based methods like MeRL improve to 74.9\%/74.8\%. Rule-based synthesis methods (Rule-SQL) are notably lower (61.1\%/61.0\%). RAT-SQL attains 74.1\%/76.3\% on WikiSQL.   Off-the-shelf SLMs perform poorly without task-specific training: Qwen-1.5B achieves 33.4\%/35.6\%, Phi-3-mini reaches 41.3\%/42.5\%, and Qwen-2.5B obtains 43.4\%/45.6\%. These results highlight that, under strict capacity constraints, directly applying compact models to Text-to-SQL yields substantial degradation.

In contrast, SQuaD-SQL substantially closes the gap between compact and large models. Using synthetic supervision generated by GPT-4o, our 1.5B-parameter student achieves 86.5\% dev and 86.9\% test accuracy, outperforming all listed Text-to-SQL baselines and recent structured QA models, and approaching GPT-4o while remaining far more resource-efficient. Overall, these results support the effectiveness of LLM-guided synthetic supervision as a practical approach for learning structured semantic parsing behaviors under resource constraints.

\begin{table}[t]
\centering
\caption{Performance comparison on the WikiSQL dataset.}
\label{tab:main_results}
\begin{tabular}{lcc}
\toprule
\textbf{Model} & \textbf{Dev Logic Accuracy} & \textbf{Test Logic Accuracy} \\
\midrule
\multicolumn{3}{l}{\textit{Large Language Models}} \\
GPT-4o~\cite{openai2024gpt4o_system_card} & 92.2\% & 93.5\% \\
\midrule
\multicolumn{3}{l}{\textit{Small Language Models (Native)}} \\
Qwen-1.5B~\cite{qwen2_2024} & 33.4\% & 35.6\% \\
Phi-3-mini~\cite{abdin2024phi3} & 41.3\% & 42.5\% \\
Qwen-2.5B~\cite{qwen2_2024} & 43.4\% & 45.6\% \\
\midrule
\multicolumn{3}{l}{\textit{Traditional Text-to-SQL Models}} \\
SQLNet~\cite{zhong2017seq2sql} & 69.8\% & 72.1\% \\
MeRL~\cite{agarwal2019learning} & 74.9\% & 74.8\% \\
Rule-SQL~\cite{guo2019towards} & 61.1\% & 61.0\% \\
RAT-SQL~\cite{wang2020rat} & 74.1\% & 76.3\% \\
MedT5SQL~\cite{marshan2024medt5sql} & 43.3\% & 44.2\% \\
TrustUQA~\cite{zhang-etal-2025-trustuqa} & 84.8\% & 85.9\% \\
M3~\cite{mouravieff-etal-2025-structural} & 80.1\% & 80.3\% \\
\midrule
\multicolumn{3}{l}{\textit{Ours (with Synthetic Data)}} \\
SQuaD-SQL (Ours) & \textbf{86.5\%} & \textbf{86.9\%} \\
\bottomrule
\end{tabular}
\end{table}

\subsection{Ablation Study}
\label{sec:ablation}

We ablate each component of SQuaD-SQL by retraining the student model under identical hyper-parameters while removing one element at a time. Results on the WikiSQL test split are reported in Table~\ref{tab:ablation}. Besides the zero-shot baseline, every variant is trained for 30 epochs on the same GPU.

\begin{table}[h!]
\centering
\caption{Ablation Studies on WikiSQL Test Data.}
\label{tab:ablation}
\begin{adjustbox}{width=\linewidth}
\begin{tabular}{ccccc}
\toprule
\textbf{Method} & \textbf{Prompt Eng.} & \textbf{Distill} & \textbf{Data Filter} & \textbf{Logic Acc. (\%)} \\
\midrule
Zero-shot & & & & 35.6 \\
\midrule
Ablation 1 & \checkmark &  & & 45.6 \\
Ablation 2 &  &\checkmark& & 80.3 \\
Ablation 3 & & \checkmark & \checkmark & 83.5 \\
Ablation 4 & \checkmark & \checkmark & & 83.2 \\
\midrule
Full (Ours) & \checkmark & \checkmark & \checkmark & \textbf{86.9} \\

\bottomrule
\end{tabular}
\end{adjustbox}

\end{table}
\vspace{1cm} 
Ablation experiments were conducted to evaluate the impact of three components of our framework: prompt engineering, LLM-based distillation, and data filtering. As shown in Tab.~\ref{tab:ablation}, using the WikiSQL-specific prompt template alone improves logical-form accuracy from the zero-shot baseline of \(35.6\,\%\) to \(45.6\,\%\), indicating the benefit of schema-aware prompting. Training solely on synthetic question–SQL pairs distilled from an LLM results in a substantial increase to \(80.3\,\%\), showing that distilled supervision is the main contributor to performance gains. Applying data filtering further improves accuracy to \(83.5\,\%\) by removing noisy examples, while combining prompt engineering with distillation achieves a similar result of \(83.2\,\%\). When all three components are combined, the model reaches \(86.9\,\%\), surpassing raw distillation by \(6.6\) points and outperforming any partial configuration. These results suggest that LLM-based distillation provides the major performance boost, while prompt engineering and data filtering offer complementary improvements that further enhance overall model quality.

\section{Conclusion}
We presented \textbf{SQuaD-SQL}, a resource-efficient framework for Text-to-SQL that bridges the performance gap between large and small language models through knowledge distillation and synthetic data generation. Our method leverages LLMs as structured knowledge sources, employing prompt engineering and rigorous data filtering to generate high-quality supervision for SLMs. This work demonstrates that SLMs, when combined with targeted LLM supervision and lightweight adaptation, can serve as practical alternatives for Text-to-SQL in resource-constrained environments.

\bibliographystyle{IEEEtran}
\bibliography{refs}
\end{document}